\documentclass[nohyperref]{article}

\usepackage{microtype}
\usepackage{graphicx}
\usepackage{subcaption}
\usepackage[export]{adjustbox}
\usepackage[font=small,labelfont=bf,tableposition=top]{caption}

\newdimen\imageheight 

\usepackage{booktabs} 

\usepackage{hyperref}

 \usepackage[accepted]{icml2022}

\usepackage{amsmath}
\usepackage{amssymb}
\usepackage{mathtools}
\usepackage{amsthm}

\usepackage[capitalize,noabbrev]{cleveref}

\theoremstyle{plain}

\theoremstyle{definition}

\theoremstyle{remark}

\usepackage[textsize=tiny]{todonotes}

\icmltitlerunning{GreenDB - A Dataset and Benchmark for Extraction of Sustainability Information of Consumer Goods}

\begin{document}

\twocolumn[
\icmltitle{GreenDB - A Dataset and Benchmark for Extraction of Sustainability Information of Consumer Goods}



\icmlsetsymbol{equal}{*}

\begin{icmlauthorlist}
\icmlauthor{Sebastian J\"ager}{equal,bht}
\icmlauthor{Alexander Flick}{equal,bht}
\icmlauthor{Jessica Adriana Sanchez Garcia}{bht}
\icmlauthor{Kaspar von den Driesch}{bht}
\icmlauthor{Karl Brendel}{bht}
\icmlauthor{Felix Biessmann}{bht,ecdf}
\end{icmlauthorlist}

\icmlaffiliation{bht}{Berliner Hochschule f\"ur Technik}
\icmlaffiliation{ecdf}{Einstein Center Digital Future, Berlin}

\icmlcorrespondingauthor{Sebastian J\"ager}{sebastian.jaeger@bht-berlin.de}

\icmlkeywords{Machine Learning, ICML}

\vskip 0.3in
]



\printAffiliationsAndNotice{\icmlEqualContribution} 

\newcommand{\code}[1]{\texttt{#1}}
\newcommand{\sebastian}[1]{\textcolor{red}{[Sebastian: #1]}}
\newcommand{\alex}[1]{\textcolor{blue}{[Alex: #1]}}
\newcommand{\felix}[1]{\textcolor{purple}{[Felix: #1]}}

\begin{abstract}
The production, shipping, usage, and disposal of consumer goods have a substantial impact on greenhouse gas emissions and the depletion of resources. 
Machine Learning (ML) can help to foster sustainable consumption patterns by accounting for sustainability aspects in product search or recommendations of modern retail platforms. However, the lack of large high quality publicly available product data with trustworthy sustainability information impedes the development of ML technology that can help to reach our sustainability goals. Here we present \emph{GreenDB}, a database that collects products from European online shops on a weekly basis. As proxy for the products' sustainability, it relies on sustainability labels, which are evaluated by experts.
The GreenDB schema extends the well-known \emph{schema.org Product} definition and can be readily integrated into existing product catalogs. 
We present initial results demonstrating that ML models trained with our data can reliably (F1 score 96\%) predict the sustainability label of products. These contributions can help to complement existing e-commerce experiences and ultimately encourage users to more sustainable consumption patterns. 
\end{abstract}

\section{Introduction}
Climate change is one of the most important challenges of our generation. 
The increasing demand for consumer goods and their implications for globalized supply chains, shipping, and disposal are major factors in increased greenhouse gas emissions~\cite{CO2_Emission_Trend, BMU_co2} and the depletion of resources. With every purchase decision made, people can decide for -- or against -- more sustainable consumption behavior. 
While many consumers would like to choose sustainable options~\cite{PWCSurvey, WWF2021, BMU_umweltbewustsein} at the point of sale, the relevant information is not available or difficult to evaluate for the consumers due to lacking standards. 
There are attempts to take sustainability information and user preferences into account for recommender systems~\cite{Tomkins20818}. Our research and discussions with researchers, managers, and engineers working for large online retailers have shown that the most important factor hindering the integration of sustainability aspects in their shopping experiences is the availability of trustworthy sustainability data. 

\begin{figure}
	\centering
	\includegraphics[width=\columnwidth, left]{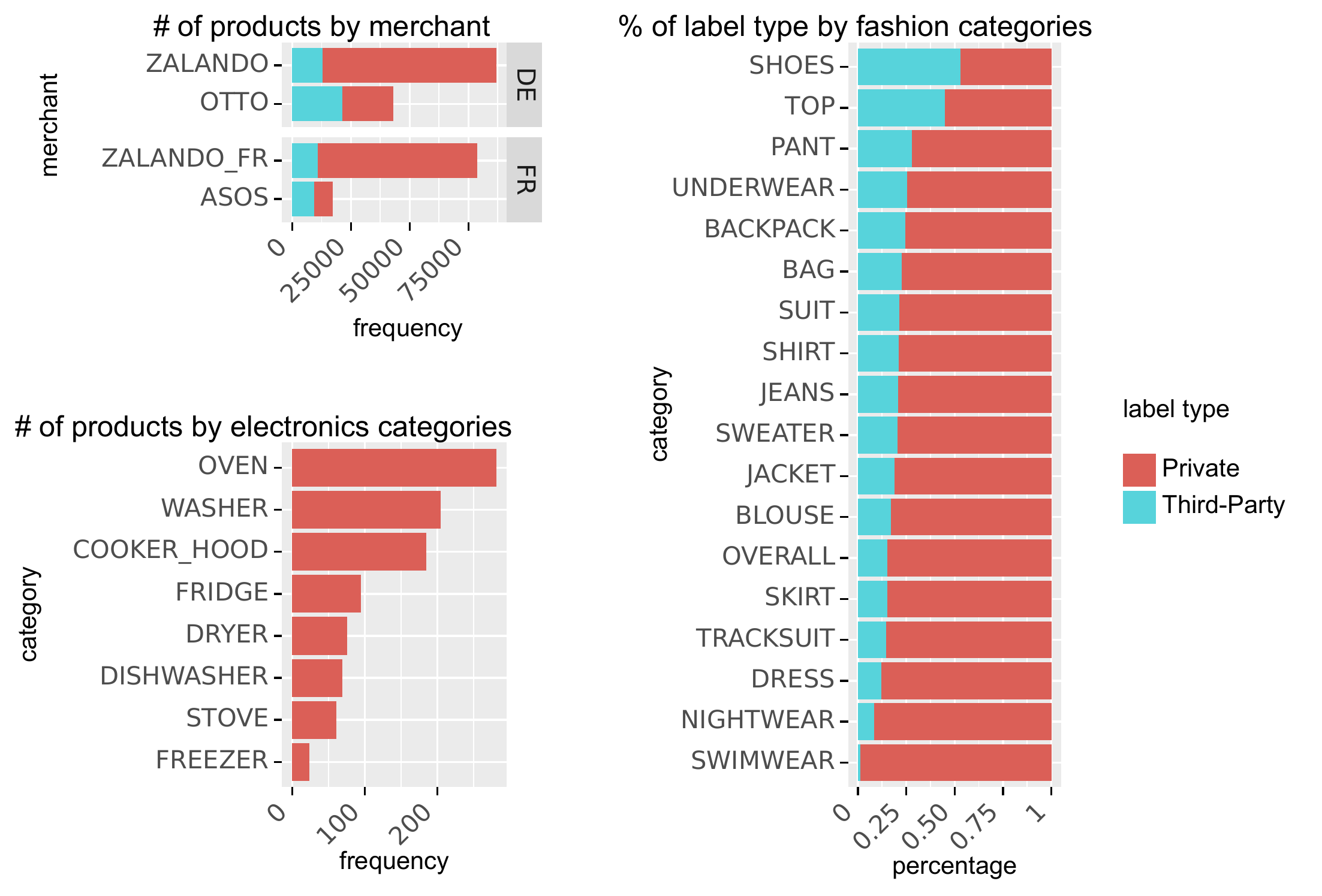}
	\caption{GreenDB overview. We follow the terminology by \cite{gossenNudgingSustainableConsumption2022} and differentiate between \emph{private} and \emph{third-party labels}. Third-party labels require independent verification of the label's environmental and/or socio-economic requirements. However, private labels are self-declared claims and not based on independent verification.
	Depending on the online shops and product categories, the number (and fraction) of preferable third-party labels varies and is generally much lower.
	}
	\label{fig:plot}
\end{figure}

To foster more sustainable shopping behavior, we need to build ML methods that are aware of sustainability. This requires large high-quality datasets that are recent, structured, and trustworthy\footnote{Many retailers and manufacturers provide sustainability information that is primarily intended for marketing purposes (greenwashing).}. 
To the best of our knowledge, the largest publicly available database containing products and their respective sustainability information is the \emph{GreenDB}~\cite{greenDB}.
As proxy for the products' sustainability, it relies on sustainability labels, which are evaluated by experts regarding their \emph{credibility}, \emph{environmental}, and \emph{socio-economic} aspects.
Currently, the GreenDB includes about $230,000$ unique products from $4$ shops in $2$ European countries. However, it  is growing continuously and is updated at a weekly cadence. Its $26$ product categories are selected based on a careful analysis of search logs of several millions of users of one of the largest search engines in Europe. An overview of the data in the GreenDB is shown in Figure \ref{fig:plot}. For detailed information about the GreenDB data collection system and sustainability evaluation process, we refer the reader to~\cite{jagerGreenDBProductbyProductSustainability2022}.

\section{Related Work}
\label{sec:related_work}
To better assess the value and contributions made with the GreenDB, we present an overview of related public datasets that complement our work.
The Web Data Commons (WDC) 
\cite{WDC} leverages structured metadata from websites HTML to create product datasets for product matching and product categorization. Based on the product corpus, they have published the WDC-25 Gold Standard for Product Categorization, consisting of $24,000$ product offers from different e-shops, which were manually assigned to 25 product categories. 
%
The Amazon ML Challenge aims at amazon browse node (product category) classification based on product metadata like \emph{name}, \emph{description}, and \emph{brand}. The dataset consists of about $2.9\:million$ products, each assigned to one of 9,919 browse nodes.~\cite{AmazonMLChallenge}
The Amazon Review Data focuses on product reviews. However, it also contains product attributes such as \emph{color} or \emph{size} and offers many product categories. Since the data spans from 1996 to 2018, the dataset size is enormous (about $34GB$).~\cite{AmazonReviewData}
The Klarna Product-Page Dataset was collected between 2018 and 2019 and is intended to benchmark representation learning algorithms on the task of web element prediction on e-commerce websites. It consists of $51,701$ product pages as HTML and screenshots of the website and has been manually labeled, specifically, the product \emph{price}, \emph{name}, \emph{image}, \emph{add-to-cart}, and \emph{go-to-cart buttons}. In addition, the data comes from 8 different markets (US, GB, SE, NL, FI, NO, DE, AT).~\cite{Klarna_Product_Page_Dataset}
Flipkart Products is most similar to GreenDB. It is a tabular dataset with $20,000$ products and 15 columns. These attributes are partially very similar to ours, e.g., \emph{product\_name}, \emph{product\_url}, \emph{retail\_price}, \emph{description}, and \emph{brand}.~\cite{FlipkartProducts}
However, none of them integrates products' sustainability information, nor are they updated regularly. 

More recently, sustainability certification institutions aim at publishing their data, such as the \emph{European Product Database for Energy Labeling (EPREL)}~\cite{EPREL}, \emph{Blue Angel}~\cite{GermanEcolabel}, or \emph{Higg Sustainability Profiles}~\cite{HiggSustainabilityProfiles}. These data sets are rather small and contain very specific selections of products.


\begin{table}
	\scriptsize
	\begin{tabular}{lrrr}
		\toprule
		{} &  precision &  recall &  f1 \\
		\midrule
		BETTER\_COTTON\_INITIATIVE         &       0.89 &    0.93 &      0.91 \\
		BIORE                            &       0.98 &    0.98 &      0.98 \\
		BLUESIGN\_APPROVED                &       0.92 &    0.92 &      0.92 \\
		BLUESIGN\_PRODUCT                 &       0.95 &    0.97 &      0.96 \\
		COTTON\_MADE\_IN\_AFRICA            &       0.95 &    0.98 &      0.97 \\
		CRADLE\_TO\_CRADLE\_BRONZE          &       0.33 &    0.14 &      0.20 \\
		CRADLE\_TO\_CRADLE\_GOLD            &       0.54 &    0.24 &      0.33 \\
		CRADLE\_TO\_CRADLE\_SILVER          &       0.75 &    0.27 &      0.40 \\
		ECOCERT                          &       0.94 &    0.68 &      0.79 \\
		EU\_ECOLABEL\_TEXTILES             &       1.00 &    0.91 &      0.95 \\
		FAIRTRADE\_COTTON                 &       0.97 &    0.85 &      0.90 \\
		FAIRTRADE\_TEXTILE\_PRODUCTION     &       1.00 &    0.80 &      0.89 \\
		GLOBAL\_RECYCLED\_STANDARD         &       0.93 &    0.85 &      0.89 \\
		GOOD\_CASHMERE\_STANDARD           &       0.00 &    0.00 &      0.00 \\
		GOTS\_MADE\_WITH\_ORGANIC\_MATERIALS &       0.84 &    0.40 &      0.54 \\
		GOTS\_ORGANIC                     &       0.85 &    0.79 &      0.82 \\
		GREEN\_BUTTON                     &       0.96 &    0.88 &      0.92 \\
		LEATHER\_WORKING\_GROUP            &       0.93 &    0.93 &      0.93 \\
		MADE\_IN\_GREEN\_OEKO\_TEX           &       0.98 &    0.95 &      0.96 \\
		ORGANIC\_CONTENT\_STANDARD\_100     &       0.79 &    0.59 &      0.67 \\
		ORGANIC\_CONTENT\_STANDARD\_BLENDED &       0.62 &    0.41 &      0.49 \\
		PRIVATE\_LABEL                            &       0.97 &    0.98 &      0.98 \\
		RECYCLED\_CLAIM\_STANDARD\_100      &       0.77 &    0.50 &      0.61 \\
		RECYCLED\_CLAIM\_STANDARD\_BLENDED  &       1.00 &    0.93 &      0.97 \\
		RESPONSIBLE\_DOWN\_STANDARD        &       0.94 &    0.72 &      0.81 \\
		RESPONSIBLE\_WOOL\_STANDARD        &       0.72 &    0.54 &      0.62 \\
		\midrule
		accuracy                                     &       0.96 &    0.96 &      0.96 \\
		macro avg                                    &       0.83 &    0.69 &      0.74 \\
		weighted avg                                 &       0.96 &    0.96 &      0.96 \\
		\bottomrule
	\end{tabular}
	\caption{Prediction results of sustainability labels based on the GreenDB product attributes \emph{name}, \emph{description}, \emph{brand}, \emph{category}, \emph{color}, \emph{merchant}, and \emph{price}. The extraction of these attributes is part of the GreenDB data collection system~\cite{jagerGreenDBProductbyProductSustainability2022}.}
	\label{tab:results}
\end{table}

\section{Predicting Sustainability Information}
To establish a simple baseline for automated inference of sustainability information from online products, we used the \emph{TabularPredictor} module of the AutoML package \emph{AutoGluon}~\cite{autogluon, agtabular}.
The model was trained on the entire GreenDB dataset~\cite{greenDB}.
We simplified the prediction problem, without loss of generality, from a multilabel problem (each product can have more than one sustainability label) to a multiclass prediction problem: For each product, we consider the first sustainability label (sorted in alphabetical order). In \autoref{tab:results}, we list the results. Note that our simple benchmark achieves an F1 score (class frequency weighted) of 0.96, indicating that  the high data quality of the GreenDB enables researchers to quickly achieve production-ready ML-based automated sustainability inference.

\scriptsize
\bibliography{greenDB-DataPerf}
\bibliographystyle{icml2022}

%

\end{document}